\documentclass{article}
\usepackage{ijcai18}

\usepackage{times}
\usepackage{xcolor}
\usepackage{soul}
\usepackage[utf8]{inputenc}
\usepackage[small]{caption}

\usepackage{helvet}  
\usepackage{courier}  
\usepackage{url}  
\usepackage{graphicx}

\def\KL{{\rm KL}}

\title{Replicating Active Appearance Model by Generator Network}

\author{
Tian Han$^1$,
Jiawen Wu$^2$,
and Ying Nian Wu$^1$
\\ 
$^1$ University of California, Los Angeles \\
$^2$ Beijing Institute of Technology \\
hantian@ucla.edu,
jiawen-wu@outlook.com, 
ywu@stat.ucla.edu
}

\begin{document}

\maketitle

\begin{abstract}
A recent Cell paper~\cite{chang2017code} reports an interesting discovery. For the face stimuli generated by a pre-trained active appearance model (AAM), the responses of neurons in the areas of the primate brain that are responsible for face recognition exhibit strong linear relationship with the shape variables and appearance variables of the AAM that generates the face stimuli. 
In this paper, we show that this behavior can be replicated by a deep generative model called the generator network, which assumes that the observed signals are generated by  latent random variables  via a top-down convolutional neural network. Specifically, we  learn the generator network from the face images generated by a pre-trained AAM model using variational auto-encoder, and we show that the inferred latent variables of the learned generator network have strong linear relationship with the shape and appearance variables of the AAM model that generates the face images. Unlike the AAM model that has an explicit shape model where the shape variables generate the control points or landmarks, the generator network has no such shape model and shape variables.  Yet the generator network can learn the shape knowledge in the sense that  some of the latent variables of the learned generator network capture the shape variations in the face images generated by AAM. 
\end{abstract}

\section{Introduction}


Recently, a paper published in Cell \cite{chang2017code} reports an interesting discovery about the neurons in the middle lateral (ML)/middle fundus (MF) and anterior medial (AM) areas of the primate brain that are responsible for face recognition. Specifically, the paper is concerned with how these neurons respond to and encode the face stimuli generated by a pre-trained  Active Appearance Model (AAM) \cite{cootes2001active,cootes2015active}.  In AAM, there are explicit shape variables and appearance variables  that generate the positions of the control points and the nominal face image respectively, and the output image is then generated by wrapping the nominal face image using the control points.  \cite{chang2017code}  discovers that the responses of the aforementioned neurons  to the face image generated by the AAM exhibits strong linear relationship with the shape and appearance variables  of the AAM that generates the face image. In fact, the shape and appearance variables of the AAM can be recovered from the neuron responses so that the face image can be reconstructed by the AAM using the recovered shape and appearance variables. 

 
In this paper, we investigate whether the above phenomenon can be replicated by deep generative models. In particular, we focus on a popular deep generative model called the  {generator network} \cite{goodfellow2014generative}, which can be considered a non-linear generalization of the factor analysis model. Recall that in the factor analysis model, the signal is generated by latent factors  that are assumed to be independent Gaussian random variables, and the signal is a linear transformation of the latent variables (plus observational noises). In the generator network, the latent variables still follow a simple known prior distribution such as independent Gaussian or uniform distribution, but the mapping from the latent variables to the observed signal is modeled by a convolutional neural network (ConvNet), which has proven to be an exceedingly powerful approximator of high-dimensional non-linear mappings. 

Both the AAM and the generator network are latent variable models where the signal is obtained by transforming the latent variables. In the AAM, the latent variables consist of explicit shape variables and appearance variables, which generate the control points and the appearance image by linear mappings learned by principal component analysis (PCA). The output image is generated by a highly non-linear but known warping function of the control points and the nominal image. In contrast, the generator network is more generic, in that it does not assume any prior knowledge about shape and deformation, and it does not have any explicit shape variables and shape model. We are interested in whether the generator model can replicate the AAM in the sense that whether the generator network can learn from the images generated by a pre-trained AAM, so that the latent variables of the learned generator network are closely related to the latent shape and appearance variables of the AAM, and the non-linear mapping from the latent variables to the output image in the generator network accounts for the highly non-linear warping function of the AAM. As it is impossible for the latent variables of the learned generator network to be the same as the latent variables of the AAM, a strong linear relationship between the two sets of latent variables (or codes) is the best we can hope for. We shall show that such a linear relationship indeed exists, thus qualitatively reproducing the behavior of the neuron responses (or neural code) observed by \cite{chang2017code}. 

The generator network can be trained by various methods, including the wake-sleep algorithm \cite{hinton1995wake}, variational autoencoder (VAE) \cite{KingmaCoRR13,RezendeICML2014,salimans2015markov}, generative adversarial networks (GAN) \cite{goodfellow2014generative,radford2015unsupervised,denton2015deep}, moment matching networks \cite{li2015generative}, alternating back-propagation (ABP) \cite{Han2017}, and other related methods \cite{oord2016pixel,Dinh2016DensityEU}. They have led to impressive results in a wide range of applications, such as image/video synthesis \cite{Alexey2015}, disentangled feature learning \cite{chen2016infogan,higgins2016beta} and pattern completion \cite{Han2017} etc. 

In this paper, we shall adopt the VAE method to train the generator network. Unlike GAN, the VAE complements the generator network with an inference network that transforms the observed image to the latent variables. The inference network seeks to approximate the posterior distribution of the latent variables given the observed image. The inference network and the generator network form an auto-encoder, where the inference network plays the role of the encoder that encodes the signal into the latent variables (or latent code), and the generator network plays the role of the decoder that decode the latent variables (or latent code) back to the signal. The parameters of the two networks can be learned by maximizing a variational lower bound of the log-likelihood \cite{blei2017variational}. We show that the latent variables computed by the inference network from the observed face image are highly correlated with the latent variables of the AAM that generates the face image. 


\noindent{\bf Contributions.} This paper is phenomenological in nature. It is our hope that the paper is of interest to both the neuroscience community and the deep learning community.  The followings are the contributions of this paper:
\begin{itemize}
\item We study the linear relationship between the latent code learned by the generator network and the AAM code that generates the face stimuli. Our experiments suggest that the deep generative model exhibits similar behavior as the primate neural system. 
\item We show that the latent variables learned by the generator network can be separated into shape-related part and appearance-related part, and the generator network is expressive enough to replicate the AAM model. 
\end{itemize}

\section{Active Appearance Model (AAM)}\label{sec:AAM} 

The active appearance model \cite{cootes2001active,cootes2015active} is a generative model for representing face images. It has a shape model and an appearance model. Both models are learned by principal component analysis (PCA). 

\textit{Shape model}: The shape model is based on a set of landmarks or control points. In the training stage, the control points are given for each training image.  Let $x$ denote the coordinates of all the control points. The shape model is 
\begin{eqnarray}
\label{eqn1}
x = \bar{x} + P_s b_s, 
\end{eqnarray}
where $\bar{x}$ is the average shape, $P_s$ is the matrix of eigenvectors, and $b_s$ is the vector of shape variables. $(\bar{x}, P_s)$ are shared across all the training examples, while $x$ and $b_s$ are different for different examples. The model can be learned from the given control points of the training images by PCA, where the number of eigenvectors is determined empirically. 

\textit{Appearance model}: The appearance model generates the nominal image before shape deformation. To learn the model, we can wrap each training image to the shape-normalized image so that its control points match those of the mean shape $\bar{x}$. Then a PCA is performed on the shape-normalized training images.  Let $g$ denote the vector of the grey-level image. The appearance model is
\begin{eqnarray}
\label{eqn2}
g = \bar{g} + P_a b_a, 
\end{eqnarray}
where $\bar{g}$ is the mean normalized grey-level image, $P_a$ is the matrix of eigenvectors,  and $b_a$ is the vector of appearance variables. $(\bar{g}, P_a)$ are shared by all the training examples, while $g$ and $b_a$ are different for different examples. 

We can learn $(P_s, P_a)$ from the training images with given control points. We concatenate the shape and appearance variables to form the face representation or the latent code, i.e., $Z_{AAM} = [b_s, b_a]$. Given $Z_{AAM}$, we can generate face image $Y$ by generating $x$ and $g$ first, and then warping $g$ according to $x$ using a warping function to output the image $Y = h(g, x)$. The warping function $h$ is given and is highly non-linear in $g$ and $x$.

\section{Generator Network} \label{sec:VAE} 

The generator network is a deep generative model of the following form: 
\begin{eqnarray}
\label{eqn3}
&&Z \sim {\rm N}(0, I_d), \\
 && Y = f_\theta(Z) + \epsilon_i, 
\end{eqnarray}
where $Z$ is the vector of latent variables (or latent code), $d$ is the dimension of $Z$, i.e., the number of latent variables. $Z$ is assumed to follow a simple prior distribution where each component is a Gaussian ${\rm N}(0, 1)$ random variable ($I_d$ is the $d$-dimensional identity matrix). The latent vector $Z$ generates the output image $Y$ by a non-linear mapping $f_\theta$, which is modeled by a top-down convolutional neural network (ConvNet), where $\theta$ collects all the weight and bias parameters of the top-down ConvNet. $\epsilon$ is the noise vector whose elements are independent ${\rm N}(0, \sigma^2)$ random variables. Even though $Z$ follows a simple distribution, the model can generate $Y$ with very complex distribution and with very rich patterns because of the expressiveness of $f_\theta$. The generator model (\ref{eqn3}) is a generalization of the factor analysis model, where the mapping from $Z$ to $Y$ is assumed linear. 

Compared to the AAM, the generator network has no explicit shape model such as (\ref{eqn1}) with control points $x$ and shape variables $b_s$, nor does it have the explicit non-linear  warping function $Y = h(g, x)$. The generator network relies on the highly expressive ConvNet $f_\theta$ to account for the linear shape model and the non-linear warping function. Even though no prior knowledge of shape and warping is built into the generator network, it can learn such knowledge by itself. 

Specifically, we shall use a pre-trained AAM as a teacher model, and we let the generator network be the student model. The AAM generates training images, and the generator network learns from the training images. We shall show that the inferred $Z$ from the face image $Y$ has a strong linear relationship with the corresponding $Z_{AAM}$ that the AAM uses to generate $Y$. 

\section{Variational auto-encoder (VAE)} 

Given a set of $N$ training images $\{Y_i, i = 1, ..., N\}$  generated by AAM, we train the generator network by variational auto-encoder (VAE) \cite{KingmaCoRR13,RezendeICML2014,salimans2015markov}. Let $p(Z)$ be the prior distribution of $Z$. Let $p_\theta(Y|Z)$ be the conditional distribution of the image $Y$ given the latent vector $Z$. Then the marginal distribution of $Y$ is $p_\theta(Y) = \int p_\theta(Z, Y) dZ = \int p(Z) p_\theta(Y|Z) dZ$. The log-likelihood is $\sum_{i=1}^{N} \log p_\theta(Y_i)$, and in principle $\theta$ can be estimated by maximizing the log-likelihood. However, this is intractable because $p_\theta(Z)$ involves intractable integral. The EM algorithm \cite{dempster1977maximum} is also impractical because the posterior distribution $p_\theta(Z|Y) = p(Z) p_\theta(Y|Z)/p_\theta(Y)$ is intractable. The basic idea of VAE is to approximate the posterior distribution $p_\theta(Z|Y)$ by a tractable inference model $q_\phi(Z|Y)$ with a separate set of parameters $\phi$, such as a Gaussian distribution with independent components ${\rm N}(\mu_\phi(Y), \sigma^2_\phi(Y))$, where $\mu_\phi(Y)$ is the vector of means of the components of $Z$, and $\sigma^2_\phi(Y)$ is the vector of variances of the components of $Z$. Both $\mu_\phi(Y)$ and $\sigma_\phi(Y)$ can be modeled by bottom-up ConvNets. 

The parameters $(\theta, \phi)$ can be learned by jointly maximizing the variational lower bound of the log-likelihood 
\begin{eqnarray}
\label{eqn4}
 L(\theta, \phi) = \sum_{i=1}^{N}\Big[ \log p_\theta(Y_i) - {\rm KL}(q_\phi(Z_i|Y_i)|| p_\theta(Z_i|Y_i)) \Big], 
\end{eqnarray}
where ${\rm KL}(q||p)$ denotes the Kullback-Leibler divergence from $q$ to $p$. $L(\theta, \phi)$ is computationally tractable as long as the inference model $q_\phi(Z|Y)$ is tractable. See \cite{kingma2013auto} for more details. $q_\phi(Z|Y)$ is the encoder, and $p_\theta(Y|Z)$ is the decoder. After learning $(\theta, \phi)$, we can estimate $Z$ from $Y$ by the learned posterior mean vector $Z_G  =  \mu_\phi(Y)$. In our work, we use $Z_G$ as the code of $Y$.


We can understand VAE as follows. Let $q_{\rm data}(Y)$ be the data distribution. Then the maximum likelihood is equivalent to minimizing ${\rm KL}(q_{\rm data}(Y)||p_\theta(Y))$ over $\theta$. VAE is equivalent to minimizing 
\begin{eqnarray}
\label{eqn5}
&&{\rm KL}(q_{\rm data}(Y)||p_\theta(Y)) + {\rm KL}(q_\phi(Z|Y)||p_\theta(Z|Y)) \nonumber \\
&=& {\rm KL}(q_{\rm data}(Y) q_\phi(Z|Y)||p(Z) p_\theta(Y|Z)) \\
&=&  {\rm KL}(q_\phi(Z, Y)||p_\theta(Z, Y)) 
\end{eqnarray}
over both $\theta$ and $\phi$, where $q_\phi(Z, Y) = q_{\rm data}(Y) q_\phi(Z|Y)$ and $p_\theta(Z, Y) =  p(Z) p_\theta(Y|Z)$.  Unlike the maximum likelihood objective function ${\rm KL}(q_{\rm data}(Y)||p_\theta(Y))$, which is the KL divergence between the marginal distributions, the variational objective function   ${\rm KL}(q_\phi(Z, Y)||p_\theta(Z, Y))$  is the KL divergence between the joint distributions. While the marginal distribution $p_\theta(Y)$ is intractable, the joint distribution $p_\theta(Z, Y)$ is tractable. 
\begin{figure}[h]
\begin{center}
\includegraphics[width=0.2\textwidth]{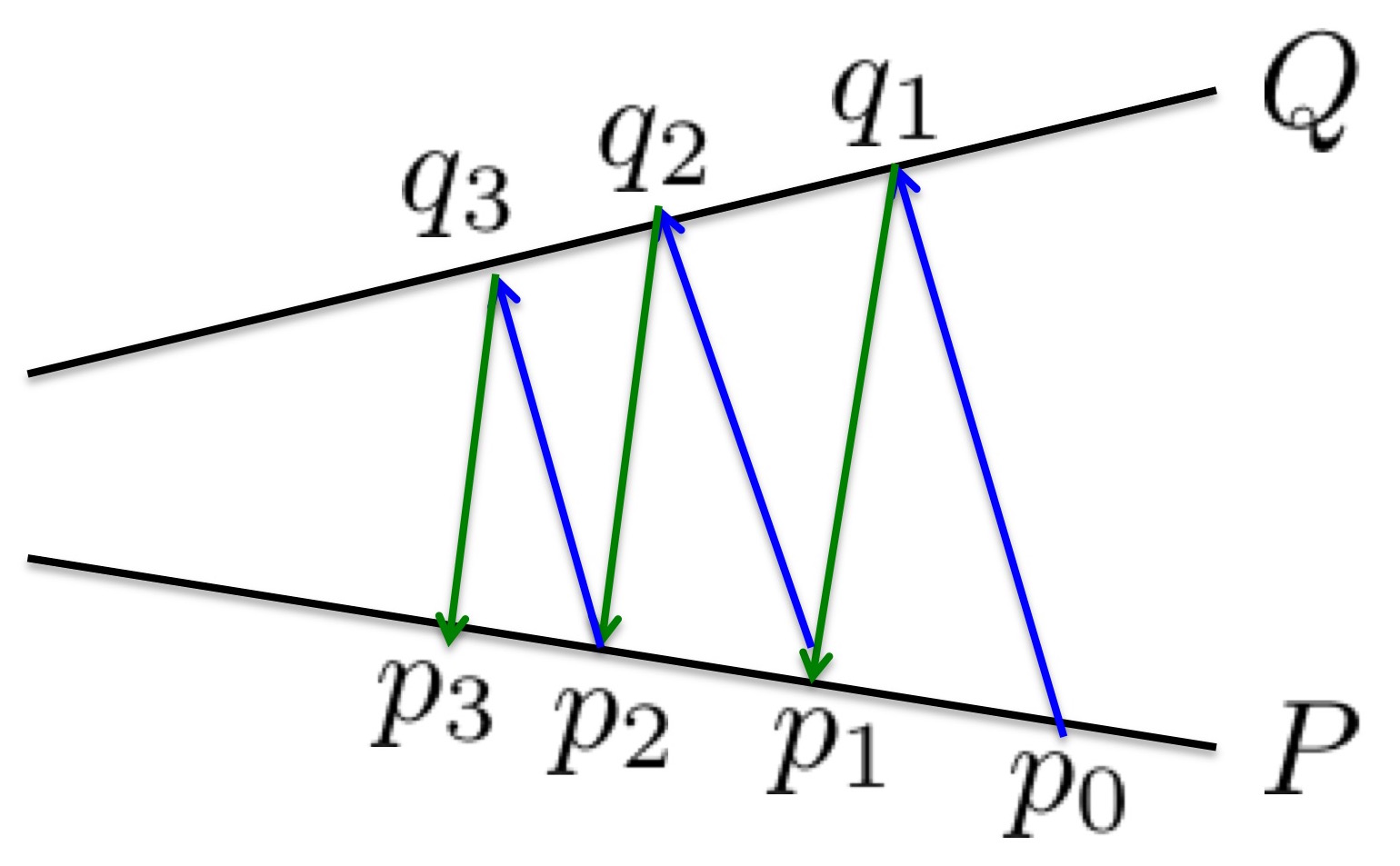}
\caption{VAE as the joint minimization of $\KL(q||p)$ over $Q$ and $P$ via alternating projection. }
\label{fig:VAE}
\end{center}
\end{figure}

Define $Q = \{ q_\phi(Z, Y) = q_{\rm data}(Y) q_\phi(Z|Y), \forall \phi\}$ and $P = \{ p_\theta(Z, Y) =  p(Z) p_\theta(Y|Z), \forall \theta\}$ be the two families of joint distributions. We can view VAE as the joint minimization of ${\rm KL}(q||p)$ over $Q$ and $P$. Such joint minimization can be accomplished by alternating projection as illustrated by Figure \ref{fig:VAE}, where $Q$ and $P$ are illustrated by two lines, and each distribution in $Q$ and $P$ is illustrated by a point. Starting from $p = p_0 \in P$, we project $p_0$ onto $Q$ by minimizing ${\rm KL}(q||p)$ over $q \in Q$ to obtain $q_1$. Then we project $q = q_1$ onto $P$ by minimizing ${\rm KL}(q||p)$ over $p \in P$ to obtain $p_1$, and so on. This process will converge to a local minimum of ${\rm KL}(q||p)$. 
In Figure \ref{fig:VAE}, the two projections are illustrated by two different colors, because they are of different natures. $\min_{q\in Q}(q|p)$ is a variational projection that minimizes over the first argument, while  $\min_{p \in P}(q|p)$ is a model fitting projection that minimizes over the second argument. As is commonly known, the former has mode seeking behavior while the latter has moment matching behavior.  

A precursor to VAE is the wake-sleep algorithm  \cite{hinton1995wake}, which amounts to replacing minimizing ${\rm KL}(q||p)$ over $q \in Q$ by minimizing ${\rm KL}(p||q)$ over $q$ by switching the order of $p$ and $q$. The minimization of ${\rm KL}(p||q)$ can be accomplished by generating data from $p$ in the sleep phase and learn $q_\phi$ from the generated data. Because of the switched order, the wake-sleep algorithm does not have a single objective function. However, in wake-sleep algorithm, both projections are of the model fitting type. 

The generator network can also be trained by GAN \cite{goodfellow2014generative}. However, GAN does not have an inference model or an encoder, which is crucial for our work. 



\section{Experiments} \label{sec:exp} 

We conduct experiments to investigate whether the generator network can replicate or imitate the AAM, where the AAM serves as the teacher model and the generator network plays the role of the student model. In the learning stage, the generator network only has access to the images generated by the AAM. It does not have access to the shape and appearance variables (latent code) used by the AAM to generate the images. After learning the generator network, we investigate the relationship between the latent code of the learned generator network and the latent code of the AAM. 

\subsection{Experiment Setting}
\noindent \textbf{Data Generation.} We pre-train the AAM using approximately 200 frontal face images with given landmarks or control points. Coordinates of the landmarks are first averaged, then PCA is performed where the first 10 principal components (PCs) for shape (see Eqn \ref{eqn1}) are retained. The landmarks of each training image are then smoothly morphed into the average shape, so that the resulting image only carries shape-free appearance information. Another PCA is then performed on the shape-normalized training images, where the first 10 PCs for appearance (see Eqn \ref{eqn2}) are retained. This results in a 20-dimensional latent face space, where every point represents a face. In other words, every face has a corresponding 20-dimensional AAM code denoted as $Z_{AAM}$, which encodes its shape and appearance variables. 

\begin{figure}[h]
	\begin{center}
		\includegraphics[width=.9\linewidth]{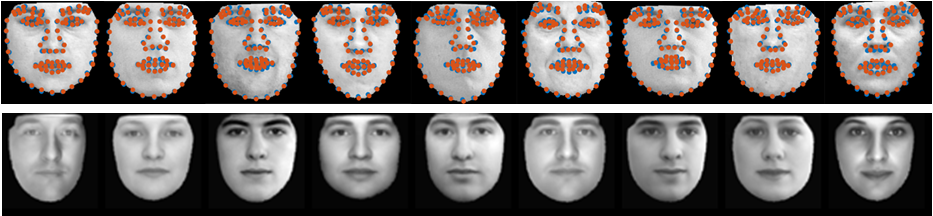}\hspace{0.5mm}			
		\caption{Top Row: training images with landmarks labeled for AAM. Bottom Row: synthesized AAM images for training the generator network.}
		\label{fig:AAM_train}
	\end{center}
	
\end{figure}

To generate face stimuli for our experiments, we randomly generate $20,000$ face images from the above pre-trained AAM. Specifically, for each dimension of the latent code, we record the standard deviation of the training responses of that dimension, and sample the variable from the Gaussian distribution with the same standard deviation as the real training faces. After that, these sampled variables are combined with the learned eigenvectors $P_s$ and $P_a$ to generate the synthesized images. The obtained images are then used as our training data for the generator network. Figure \ref{fig:AAM_train} shows some examples of training images to pre-train the AAM, and the synthesized face images generated by the trained AAM. 

\noindent \textbf{VAE Training.} The training images obtained above are scaled so that the intensities are within the range $[-1,1]$. The training images are also re-sized to $[64, 64]$ to ease the computation. 

For the generator network, we adopt the structure similar to \cite{radford2015unsupervised,Alexey2015}. The network consists of multiple deconvolution (a.k.a convolution-transpose) layers interleaved with ReLU non-linearity and batch normalization \cite{Ioffe2015BatchNA}. Specifically, we learn a 5 layer top-down convNet. The first deconvolutional layer has $512$ filters with kernel size $4\times4$ and stride  $1$. There are $256, 128, 64, 1$ filters with kernel size $4\times4$ and stride $2$ for the second, third, fourth and fifth deconvolution layers respectively. Each deconvolution layer is followed by ReLU non-linearity and batch normalization except the last deconvolution layer which is instead followed by the tanh non-linearity. 

For the inference model or the encoder network of VAE, we utilize the mirror structure of the generator network (which is the decoder network) where we use convolutional layers instead of deconvolutional ones. Besides, we use the ReLU with leaky factor $0.2$ as our non-linearity. The mean and variance networks of the inference model share the same network structure except the top fully-connected layer. We also adopt the batch normalization in the inference model as in the generator network. 

\begin{figure}[h]
	\begin{center}
		\includegraphics[width=.9\linewidth]{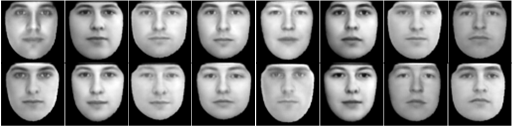} 
		\caption{Synthesized images generated by the trained generator network.}
		\label{fig:VAE_syn}
	\end{center}	
\end{figure}

We tried different dimensionalities for the latent code $Z$, including $20$, $100$ and $200$ dimensions. We used Adam optimizer \cite{kingma2014adam} with initial learning rate $0.0002$ for $500$ iterations. The outputs of the mean network of the inference model are used as the learned latent code and are denoted as $Z_{G}$.  Realistic synthesized images can be generated by the trained generator network.  See Figure \ref{fig:VAE_syn} for some examples. 

We design four experiments to examine the relationship between the AAM code $Z_{AAM}$ for generating the face images and the code learned by the generator network, $Z_{G}$. 

\subsection{Linear Relationship}
 \cite{chang2017code} discovered that  if a neuron has ramp-shaped tuning to different facial features, then its neural response can be approximated by a linear combination of the facial features. That is, the neural code for face patches ML/MF and AM has linear relationship with the AAM code of the presented face stimuli. In our first experiment, we check the strength of linearity between the code learned by the generator network and the underlying AAM code. 

The $20,000$ codes for AAM, i.e., $Z_{AAM}$, are used to predict the corresponding $20,000$ codes learned by generator network, i.e., $Z_{G}$, and vice versa. Specifically, we fit linear model A and linear model B respectively: 
\begin{eqnarray}
&& Z_{G} \approx A Z_{AAM},\\
&& Z_{AAM} \approx B Z_{G}. 
\end{eqnarray}
We also include interception terms in both models. The goodness of fit of the model is determined by the percentage of variance in data that is explained by the fitted linear model, i.e., the so-called R-square ($R^2$):
\begin{equation}
R^2 = 1 - \frac{\sum_i \|Z_i - \hat{Z_i}\|^2}{\sum_i \|Z_i - \bar{Z}\|^2},
\end{equation} 
where $Z_i$ is the given code for image $i$, $\hat{Z_i}$ denotes the fitted value, and $\bar{Z}$ is the average of the code. Higher $R^2$ indicates  stronger linear strength.

The $R^2$ values for different dimensionalities of $Z_{G}$ are shown in the first two rows of Table \ref{tab:R2}. In addition to the convolutional-based (Conv) structures of the generator network stated above, we also test the linear relationship using fully-connected (FC) structures. Specifically, we learn 4 FC layers with hidden dimension $256$, using ReLU nonlinearity for the decoder and Leaky ReLU with factor $0.2$ for the encoder, and all the layers are followed by batch normalization except the last ones. The $R^2$ values are reported in the last two rows of Table \ref{tab:R2}. It can be seen that both models show  strong linear relations. This is non-trivial and surprising, because when presented with only synthesized face stimuli, the VAE training of the highly non-linear generator network \cite{montufar2014number} can automatically learn the code that is linearly related to the underlying AAM code that generates the given face stimuli. That is, the learned generator network shares similar behavior as the face patch systems ML/MF and AM  in the primate brain. 

\begin{table}[h]
	\begin{center}
		\begin{tabular}{|c|c|c|c|}
			\hline dimension d for $Z_{G}$	&	d=20	& d=100	& d=200	\\
			\hline $R^2$ (A)(Conv)	&	0.9602	& 0.9624	& 	0.9631\\
			\hline $R^2$ (B)(Conv)    &   0.9644   &  0.9807      &   0.9889   \\ \hline
			\hline $R^2$ (A)(FC)	&	0.9585	& 0.9588	& 	0.9594\\
			\hline $R^2$ (B)(FC)   &   0.9410   &  0.9649      &   0.9709   \\
			\hline
		\end{tabular}		
		\caption{Strength of linearity ($R^2$) for models A and B with different model structures of the generator networks. } 
		\label{tab:R2}
	\end{center}
\end{table}

\subsection{Decoding}
As argued in \cite{chang2017code}, we should be able to linearly decode the facial features from the neural responses if there is a linear relationship between them. If so, we can accurately predict what the primate brain sees by knowing only the neural responses of the brain. Knowing that our learned code of the generator network $Z_{G}$ shows strong linear relationship with the facial features $Z_{AAM}$ from the above experiment, we expect that our automatically learned code $Z_G$ can accurately predict the facial features $Z_{AAM}$, which can then be used to reconstruct the input face image via the AAM. Therefore we further examine the decoding quality in this section. 

To proceed, for training, we use $20,000$ $Z_{AAM}$ and $Z_{G}$ obtained during  the learning process to fit model B as described above. Denote the estimated coefficients as $B^\star$. To test the decoding quality, we carry out the following two steps: (1) randomly sample a new set of $2,000$ AAM generated face images, which are used as the testing set. Then use the trained encoder network, i.e., mean network, of VAE to get point estimate of latent code of the generator network, i.e., $Z^{test}_{G}$. (2) Use the optimal $B^\star$ to get the predicted AAM code: 
\begin{equation}
\hat{Z}^{test}_{AAM} = B^\star Z^{test}_{G}.
\end{equation}
The predicted AAM code is then projected onto the previously  learned AAM eigenvectors $P_s$ and $P_a$ to get the reconstructed image. 

\begin{figure}[h]
	\begin{center}
		\includegraphics[width=.49\linewidth]{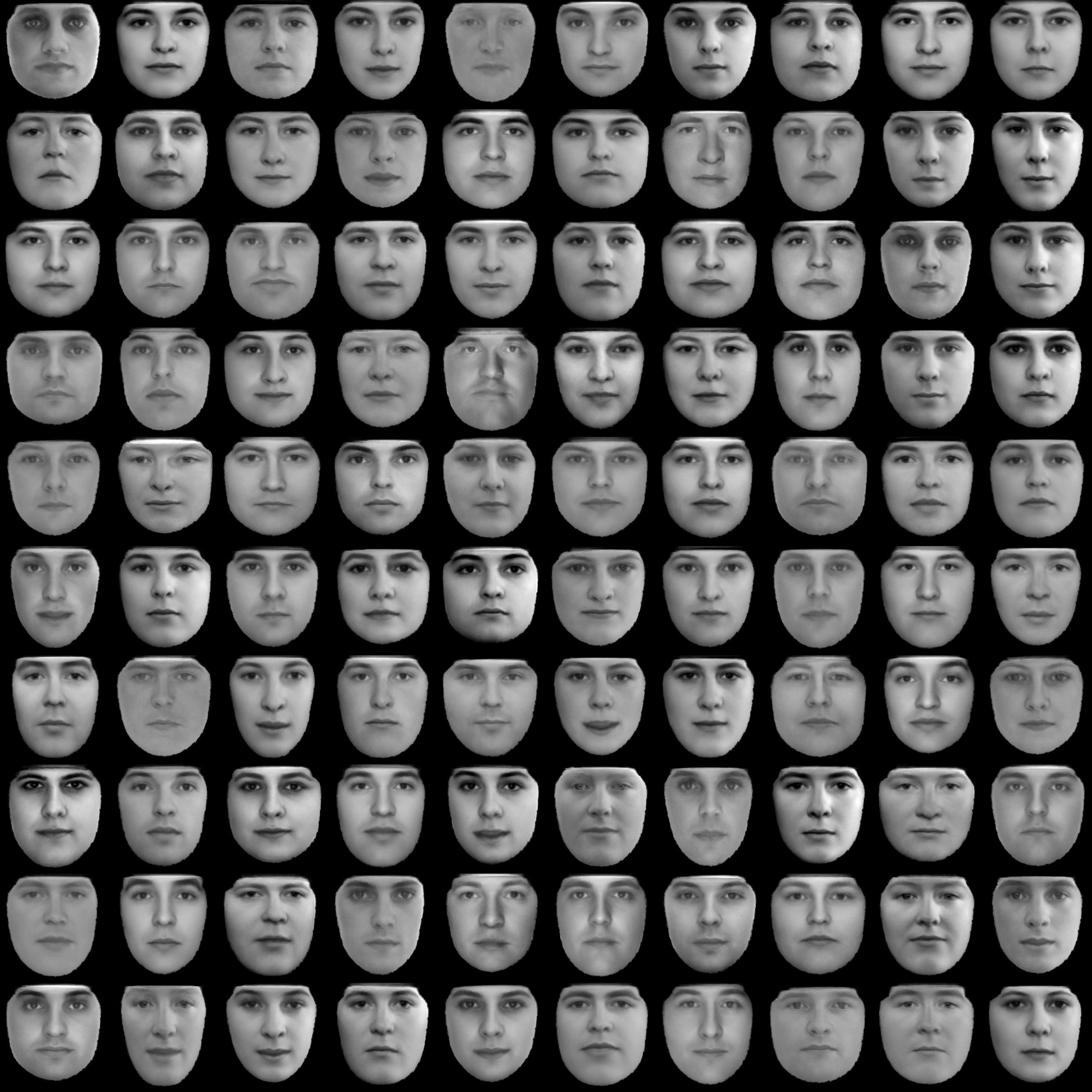}\hspace{0.5mm}
		\includegraphics[width=.49\linewidth]{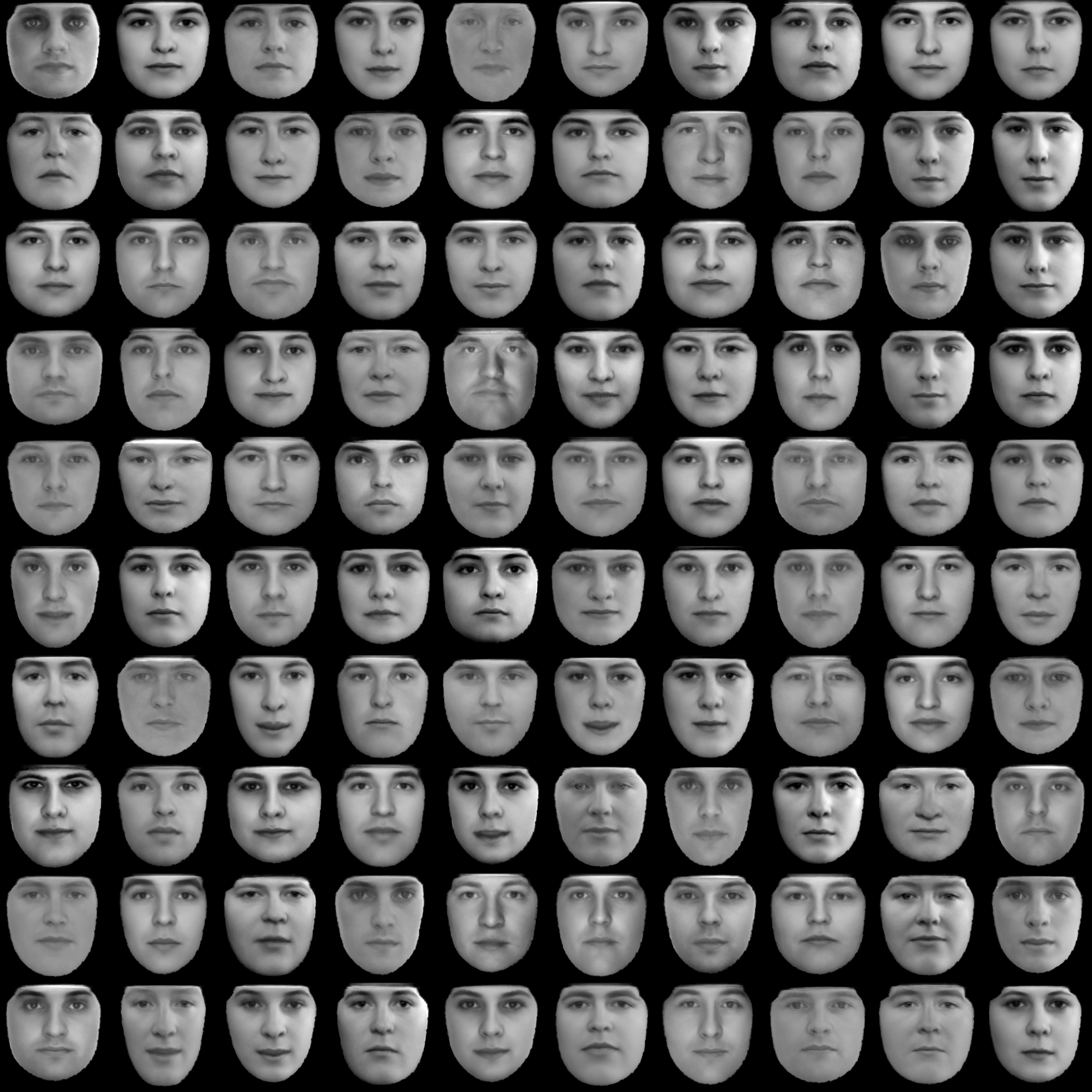}		
		\caption{Left: test faces. Right: reconstructed faces using linear decoding.}
		\label{fig:AAM}
	\end{center}
\end{figure}

Figure~\ref{fig:AAM} shows some testing images and  the reconstructed ones. It can be seen that the linear model between the learned code by VAE and the AAM code  gives us high decoding quality. In this experiment as well as the subsequent experiments, we set the dimensionality of $Z_G$ to be $100$. Other dimensionalities give similar results. 

\subsection{Shape/Appearance Separation}
The  latent code $Z_{G}$ learned by the deep generative model is  mixed with shape and appearance information.  It would be useful to separate the shape and appearance parts of $Z_{G}$. In this experiment, we further identify the strengths of shape and appearance parts of the learned code $Z_{G}$. 

From the first two experiments, we show that $Z_{G}$ is linearly related to the AAM code $Z_{AAM}$, which contains both the shape and appearance parts. We can identify these two parts by projecting each dimension of $Z_{G}$ onto the $10$ dimensional shape code $b_s$ and the $10$ dimensional appearance code $b_a$. We can then obtain the relative $R^2$ for each part. A higher $R^2$ for one part indicates the stronger response for this part. Recall that $Z_{AAM} = [b_s, b_a]$. We fit the linear models on the shape part $b_s$ and the appearance part  $b_a$ respectively:
\begin{eqnarray}
Z_{G} &\approx& A_{s}  b_s, \\
Z_{G} &\approx& A_{a}  b_a.
\end{eqnarray}     

\begin{figure}[h]
	\begin{center}
		\includegraphics[width=.8\linewidth]{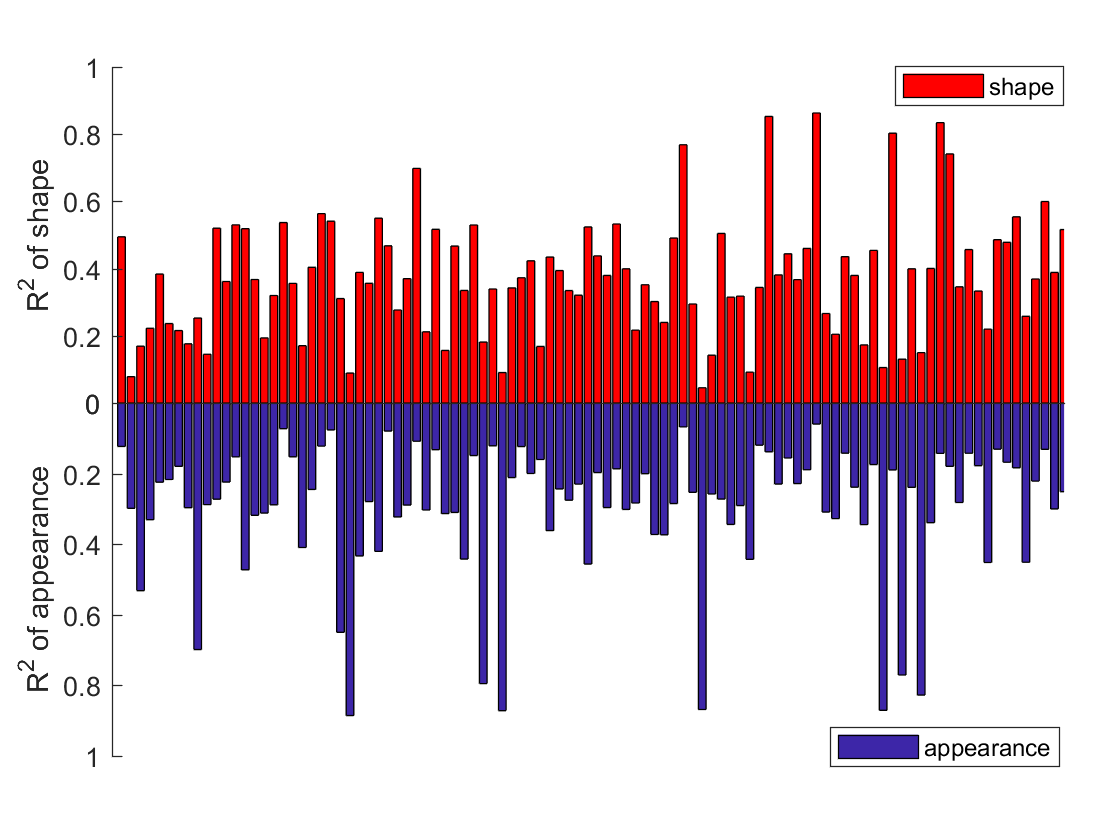}
		\caption{Bar plot for $R^2$ values of shape and appearance. Each vertical bar corresponds to a dimension of the latent code. }
		\label{fig:barSA}
	\end{center}
\end{figure}

\begin{figure}[h]
	\begin{center}
		\includegraphics[width=.9\linewidth]{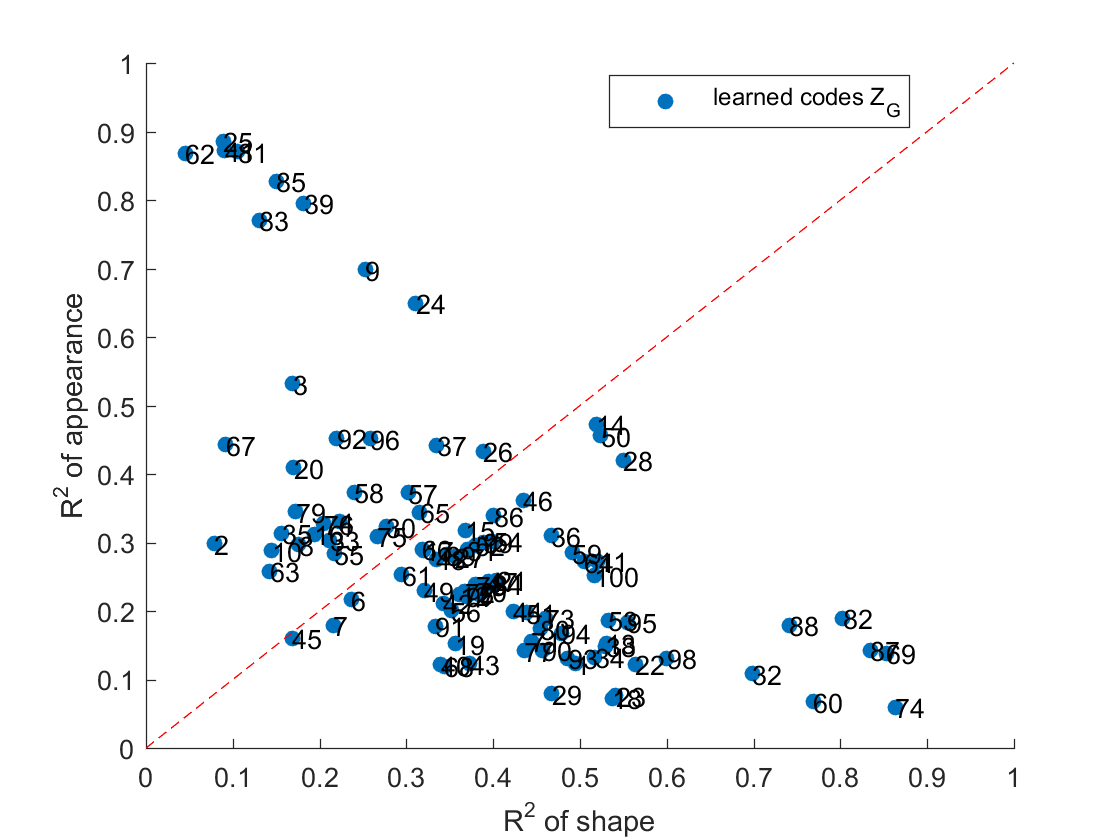}		
		\caption{Scatter plot for $R^2$ values of shape and appearance. Each point corresponds to a dimension of the latent code. Red dashed line indicates the equal $R^2$ values for shape and appearance. }
		\label{fig:scatterSA}
	\end{center}
\end{figure}

\begin{figure}[h]
	\begin{center}
		\includegraphics[width=.8\linewidth]{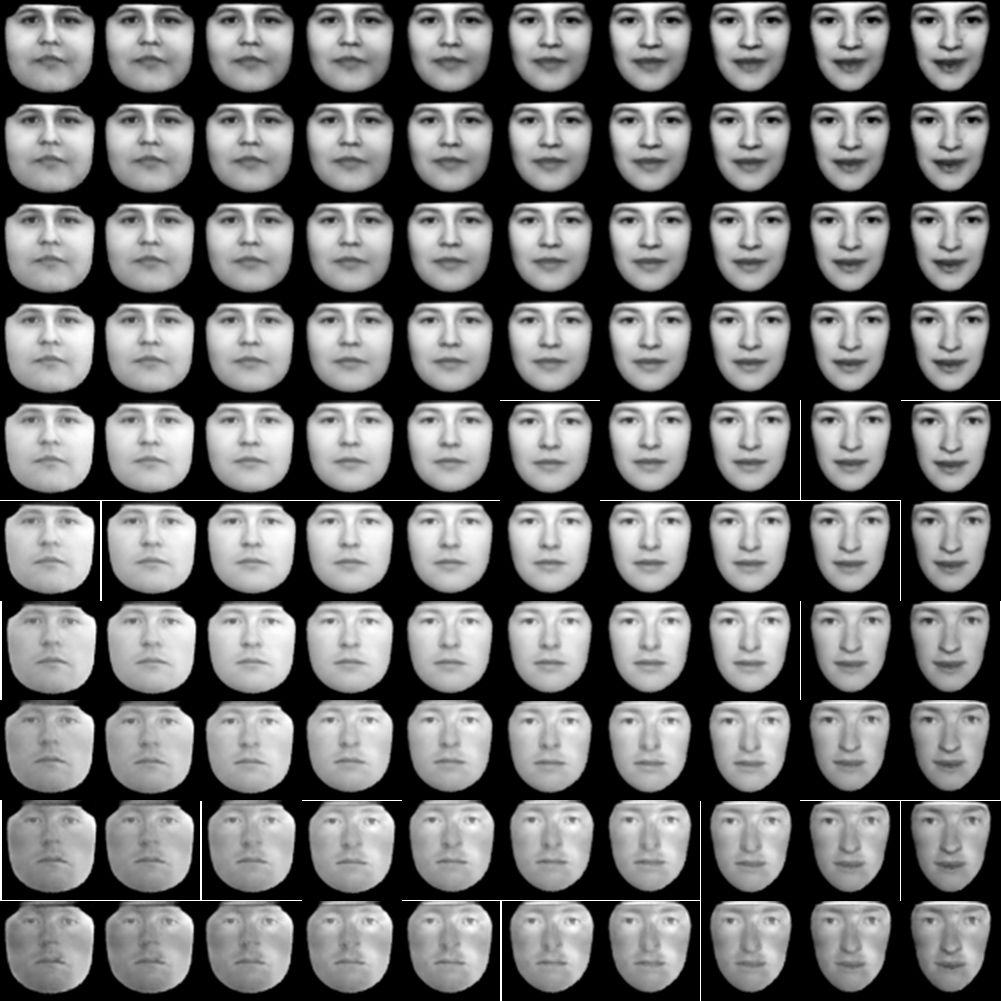}		
		\caption{Vertical: appearance variation. Horizontal: shape variation.}
		\label{fig:AAM_variation}
	\end{center}
\end{figure}

Figures \ref{fig:barSA} and \ref{fig:scatterSA} show the $R^2$ values for each dimension of $Z_{G}$. It shows that each dimension of $Z_G$ responds differently to shape and appearance. To further verify and visualize our analysis, we first choose three dimensions with the top $R^2$ for shape  and three dimensions with the top $R^2$ for appearance. Then we visualize the generated images using the trained generator network by varying ($\pm 3$ sd) the three chosen dimensions of the learned code while keeping the other three dimensions fixed. Figure \ref{fig:AAM_variation} shows the result. It is clear that if we only vary the shape dimensions of the code (horizontally in the figure), the generated images mainly change their shapes while the appearances tend to remain similar. On the other hand, if we only vary the appearance dimensions of the code (vertically in the figure), the generated images mainly change their appearances instead of shapes.  

\subsection{Replicating AAM by Supervised Learning}

So far the generator network learns from the AAM in the unsupervised manner, where the generator network only has access to the training images but not the latent code of the AAM. 
 We now examine whether the generator network has enough expressive power to replicate the AAM in the supervised setting where we also provide the latent code of the AAM to the generator. 

In this experiment, the $20,000$ synthesized face images and their AAM codes are given, and we use these pairs to learn the generator network. To be more specific, we first train the generator network using the $20,000$ images and their codes. Let us denote the trained generator as $G^\star$. Then, we prepare a new set of $2,000$ synthesized images $Y_{test}$ and their AAM code $Z^{test}_{AAM}$ as our testing set. $Z^{test}_{AAM}$ is then fed into $G^\star$ to get $\hat{Y}_{test}$. If the generator network is capable of replicating the AAM, then $\hat{Y}_{test}$ should be close to $Y_{test}$, that is, the generated images by the trained generator network should be similar to the testing face images. 

\begin{figure}[h]
	\begin{center}
		\includegraphics[width=.49\linewidth]{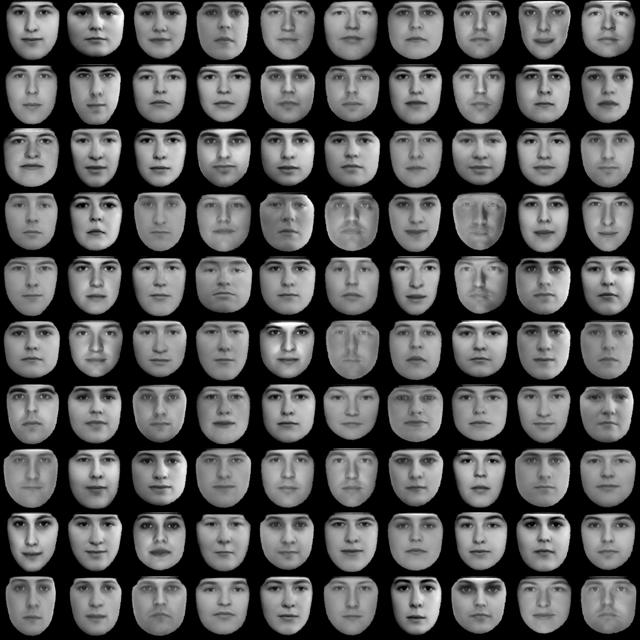}\hspace{0.5mm}
		\includegraphics[width=.49\linewidth]{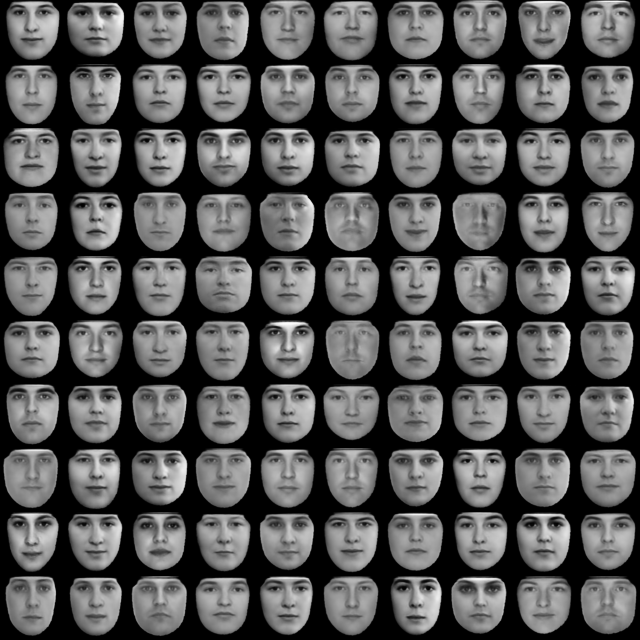}		
		\caption{Left: test face images generated by AAM. Right: reconstructed face images by the generator network trained by supervised learning.}
		\label{fig:AAM_rep}
	\end{center}
\end{figure}

We use the same generator network structure as in the VAE training. We use the stochastic gradient descent (SGD) algorithm with momentum $0.5$ to train the generator network for supervised learning. The learning rate is $0.0001$ with $900$ epochs. Figure \ref{fig:AAM_rep} shows the ground-truth testing images generated by the AAM and the reconstructed images generated by the trained generator network. We also calculate the per-pixel $\ell_1$ reconstruction error, which is 0.0113.

\section{Conclusion}
The recent work in neuroscience \cite{chang2017code} shows that the face images can be  reconstructed using the cell responses from face patches ML/MF and AM. To investigate whether the widely used generator network has the similar property, we design and conduct experiments to examine the relationship between  the AAM code that generates the face stimuli and the automatically learned code by the generator network. Through the linearity analysis and the decoding quality analysis, we  find that the biological observations made in \cite{chang2017code} can be qualitatively reproduced by the generator network, i.e., the learned code shows a strong linear relationship with the AAM code. Additionally, we can also use this relationship to further separate the shape and appearance parts of the learned code. Again this is similar to the neural system as it is found that ML/MF and AM carry complementary information about shape and appearance. 
Furthermore, we show that the generator network is capable of replicating AAM and we demonstrate this through supervised learning. 

In this paper, we distill the knowledge of a pre-trained AAM to the generator network. It will also be interesting to distill the knowledge of a learned generator network to an AAM in order to interpret the generator network. We leave it to future work.

\section*{Acknowledgments}

The work is supported by DARPA SIMPLEX N66001-15-C-4035,  ONR MURI N00014-16-1-2007, DARPA ARO W911NF-16-1-0579, and DARPA  N66001-17-2-4029. Part of the work was done while the first author was visiting Microsoft Research in Seattle. We thank Dr. Gang Hua for his help. 

\bibliographystyle{named}
\bibliography{ijcai18}

\end{document}